\documentclass{article}

\usepackage{arxiv}

\usepackage{times}

\usepackage[numbers]{natbib}
\usepackage{multicol}
\usepackage{graphicx}
\usepackage{subcaption}
\usepackage{hyperref}
\usepackage{algorithm}
\usepackage{algorithmic}
\usepackage{amsmath}
\usepackage{amsfonts}
\usepackage{amssymb}
\usepackage{placeins}
\newtheorem{theorem}{Theorem}

\newenvironment{proof}[1][Proof]{\begin{trivlist}
		\item[\hskip \labelsep {\bfseries #1}]}{\end{trivlist}}

\title{Asynchronous Deep Model Reference Adaptive Control}

\author{
   Girish Joshi$^*$\\
   \texttt{girishj2@illinois.edu} \\
   \And
   Jasvir Virdi$^{**}$\\
   \texttt{jvirdi2@illinois.edu} \\
   \And
   Girish Chowdhary$^\dagger$\\
   \texttt{girishc@illinois.edu} \\
   \\
   $^*$Department of Aerospace Engineering\\
  $^{**}$Department of Mechanical Science and Engineering \\
  $^\dagger$Department of Computer Science \& Department of Agricultural and Biological Engineering\\
  University of Illinois at Urbana-Champaign,  
  United States.
}

\begin{document}
\maketitle
%===============================================================================
\begin{abstract}
In this paper, we present Asynchronous implementation of Deep Neural Network-based Model Reference Adaptive Control (DMRAC). We evaluate this new neuro-adaptive control architecture through flight tests on a small quadcopter. We demonstrate that a single DMRAC controller can handle significant nonlinearities due to severe system faults and deliberate wind disturbances while executing high-bandwidth attitude control. We also show that the architecture has long-term learning abilities across different flight regimes, and can generalize to fly different flight trajectories than those on which it was trained. These results demonstrating the efficacy of this architecture for high bandwidth closed-loop attitude control of unstable and nonlinear robots operating in adverse situations. To achieve these results, we designed a software+communication architecture to ensure online real-time inference of the deep network on a high-bandwidth computation-limited platform. We expect that this architecture will benefit other deep learning in the closed-loop experiments on robots.

%This generalizability and long-term learning is in stark contrast with prevalent quadrotor attitude controllers which would have needed several different controllers/networks to handle each different case. %In addition, the ability to learn features directly from data afforded by DMRAC can remove a lot of the guess work in designing the features of the learning element in the adaptive control. Finally, DMRAC comes with stability guarantees, making it a very promising architecture for control of highly dynamic mobile robots. % architecture utilizes the power of deep neural network representations for modeling significant nonlinearities while marrying it with the boundedness guarantees that characterize MRAC based controllers. We demonstrate through simulations and analysis that DMRAC can subsume previously studied learning based MRAC methods, such as concurrent learning and GP-MRAC. This makes DMRAC a highly powerful architecture for high-performance control of nonlinear systems with long-term learning properties. s we test against, and suffer from forgetting from one learned test case to another. 
\end{abstract}


%===============================================================================
\section{Introduction}
Learning to control mobile robots online on computation limited platforms while ensuring that the learning transients do not destabilize the system in the presence varied disturbances and operating environments is a challenging problem. The key challenge is that robot dynamics or environmental conditions can significantly change during operation. For example, for a flying robot, uncertainty can appear in environment variables, payloads, system degradation, failures and so on. 
When such changes happen, heuristic, hand-crafted, or model-based controllers fail to achieve the control objective. Learning based controllers hold the promise to be more robust. However, traditional methods such as reinforcement learning (RL) can fail to produce sensible behavior on real robots, especially when the robot dynamics changes beyond what the RL agent was trained on. %, which has led to significant interest in the development of several learning based and adaptive control techniques for mobile robots. 
On the other hand, traditional adaptive controllers are  designed to change their behavior \textit{online and in real-time} to adapt to changes while guaranteeing system stability. Such controllers have been studied extensively in the controls community \cite{ioannou1988theory,tao2003adaptive}, however, a key problem with these traditional methods is the lack of long-term learning. More recently, Model Reference Adaptive Controllers (MRAC) using shallow networks have demonstrated some  long term learning \cite{Chowdhary13_TNN,grande:JAIS:2014}, however, the results have been limited to mild changes in the dynamics and in general limited by the learning ability of shallow networks. Yet, their capability to adapt, stability guarantees, and computational efficiency  have enabled these methods to emerge as the leading method for learning-based flight control, including for highly unstable rotorcraft with a number of real-world demonstrations \cite{kannan:jgcd:05,Chowdhary2014IJC,grande:JAIS:2014,gregory2009l1,chowdhary2013guidance}. %Methods such as Gaussian Process Model Reference Adaptive Control \cite{Chowdhary13_TNN,grande:JAIS:2014}, L1 adaptive control \cite{michini2009l1}, and single hidden layer neural network based adaptive control \cite{kannan:jgcd:05,Chowdhary:JGCD:10} have demonstrated quite a bit of success in adapting to disturbances during flight. 
A key drawback of these existing methods has been the lack of long-term learning: The shallow networks in the above mentioned methods produce a policy which optimize a Lyapunov loss function and hence can instantaneously adapt to mitigate the disturbance, but do not generalize to similar disturbances or operating conditions \cite{Chowdhary:JGCD:10,Chowdhary2014IJC,Chowdhary13_TNN}. Deep Neural Networks (DNNs) trained with dropouts and batch updates could potentially help alleviate these short-term learning issues \cite{joshi2019deep}, but it has been difficult to train and update these networks in real-time on aerial robots with limited onboard computing while guaranteeing stability in presence of significant faults.
% \begin{figure}[tbh]
%     \centering
%     \includegraphics[width=\columnwidth]{Fig/drone_with_chipped_rotor.jpg}
%     \caption{We present flight tests with a deep learning based adaptive controller on a small quadrotor subject to significant unforeseen disturbances, including blades that break in the middle of flight. Our new asynchronous fast-slow loop based adaptive control architecture implemented on an aerial robot with very limited onboard compute has desirable long-term learning and disturbance rejection properties compared to  shallow-learning based adaptive controllers.}
%     \vspace{-0.2in}
%     \label{fig:quad}
% \end{figure}
\subsection{Contributions and Significance}
Our main contribution in this paper are algorithms and a hardware-software architecture for a real-time updated deep learning based adaptive flight controller for dynamic robots. The system is implemented on a small quadrotor (Parrot Mambo Fly) with limited onboard compute. The system utilizes Deep MRAC algorithm, with contribution here being a novel fast-slow asynchronous update framework where the last layer of the network is updated online on the embedded controller on-board the aerial robot, and the deeper layers of the network are updated on a separate computer using batch updates and dropouts. Both the fast update computing on-board embedded computer, and the batch update computing off-board controllers communicate with each other in an intermittent (asynchronous) fashion, with the off-board computer intermittently sending updated inner layer parameters to the on-board computer and can tolerate significant communication dropouts.

The major significance of our contributions is that we demonstrate that when implemented using our fast-slow update strategy, deep learning based controllers hold tremendous promise for high-bandwidth adaptive control of agile robots with limited onboard compute. Our results show not only a fully-connected deep network can be included in the closed loop and updated asynchronously in a way such that the real-time flight control is possible, but also that a control framework that utilizes DNNs can adapt to handle significant disturbances such as wind bias and damage. We expect that our system can be adopted for other deep learning based controllers, including reinforcement learning and MPC, thereby enabling the incorporation of deep learning for robot control. 

\subsection{Background: Attitude Control of Quadrotor Aerial Robots}
Attitude control of aerial robots is a challenging problem due to the highly nonlinear, unstable, and fast dynamics \cite{stevens2015aircraft}. Attitude controllers require fast update rates (typically over 100~Hz), furthermore, adaptation is necessary because the flight dynamics can significantly change due to degradation, change in operating conditions, or failures \cite{Joh:02CP,Steinberg:AST:05}.  Due to the unstable dynamics and highly variable operating conditions, learning based control has been hard for quadrotors. One body of methods that have had some success is  Model Reference Adaptive Control (MRAC) \cite{kannan:jgcd:05,Chowdhary2014IJC,Joh:02CP,Steinberg:AST:05,michini2009l1,Patel:IJC:09,Lavertsky:ACC05}. %MRAC seeks to learn a high-performance control policy in the presence of significant model uncertainties that can also rapidly adapt to disturbances \cite{ioannou1988theory,tao2003adaptive, Pomet_92_TAC}. 
The key idea in MRAC is to augment a baseline (linear) flight controller, with a learning element that is updated online to improve tracking performance while ensuring Lyapunov stability. The benefit of this approach is that Lyapunov stability can be proven, and the learning problem can be extended to policy gradient like updated with a Lyapunov constraint. However, the stability results have only held for shallow learning elements, which are limited in their generalization and learning capabilities. Unlike their shallow counterparts  deep networks learn features by learning the weights of nonlinear compositions of weighted features arranged in a directed acyclic graph \cite{2013arXiv1301.3605Y}. This ability to learn features can help deep networks learn and generalize across similar examples in supervised learning problems \cite{hinton2012deep}. However, recent success of deep learning has required significant amount of training with large data on powerful computers. This has made it hard for methods that use deep learning to be implemented for control of robots where learning has to be online, in real-time, and on low-power embedded computers on small robots. %Even small GPUs like the NVIDIA Jetson Nano are not always easy to be embedded on small mobile robots due to power requirements and overheating problems. 
This is one reason real-world demonstrations controllers with DNNs, such as deep RL or neuro adaptive controllers have been largely missing, especially on challenging systems like attitude control of quadrotors.  Specifically, there have been methods in which Deep Reinforcement Learning (DRL) has been used to generate trajectories when the inner loop attitude control was achieved with PIDs \cite{lupashin2010simple,abbeel2007application,abbeel2010autonomous,kim2004autonomous}. A notable early result was Ng. et al's (non deep) RL in the outerloop for inverted flight of a rotorcraft \cite{ng2006autonomous} with the inner loop attitude control executed with LQR. However, such learning has not generalized to learning to %outerloop trajectory and velocity control is a rather different problem than 
control the much faster and unstable attitude loop on real aircraft \cite{bagnell2001autonomous}. %On a quadrotor, and for that matter on any rotorcraft aerial robot, the attitude loop is highly unstable and control commands must update in real-time at 80Hz or above \gXX{check}. This makes the application of DRL very difficult for attitude control on real aircraft. 
%Hence, even though there have been some successes in simulation on rotorcraft \cite{bagnell2001autonomous} and other continuous and high-bandwidth platforms \cite{lillicrap2015continuous}, these methods seldom make it to real robots due to computational challenges. 
Ideas of separately training different layers of the network asynchronously have been used in computer vision \cite{song2014deep}, however, not yet for closed-loop attitude control. The goal of this paper is to alleviate this bottleneck by creating a system that enables the use of deep learning based controllers on low-power robots with fast unstable dynamics.% \gXX{others}.  %There have been some successes in simulation,  but not yet real world demonstrations %. For example, %Deep reinforcement learning and other deep learning based techniques hold great promise for learning  control policies for mobile robots such as quadrotors 
%several techniques have been demonstrated in simulation for attitude control of dynamic and unstable platforms like quadrotors, but not yet in the real world 
.  %, the demonstrations have been largely limited to simulations. The physical realization of such learning based methods on a real

\section{Flight Control System Description}
\label{system_description}
\subsection{Flight control problem formulation}
%This section discusses the formulation of model reference adaptive control (see e.g. \cite{ioannou1988theory}). 
The attitude control system of a quadrotor subject to unknown disturbances can be abstracted as the below with the nonlinear and unknown  function $\Delta(x)$:
\begin{equation}
\dot x(t) = Ax(t) + B(u(t) + \Delta(x)).
\label{eq:0}
\end{equation}
where $x(t) \in \mathbb{R}^n$ is the state vector, $u(t) \in \mathbb{R}^m$, $\forall t \geqslant 0$ is the control input, $A \in \mathbb{R}^{n \times n}$, $B \in \mathbb{R}^{n \times m}$ are known system matrices and we assume the pair $(A,B)$ is controllable. The term $\Delta(x) : \mathbb{R}^n \to \mathbb{R}^m$ is matched system uncertainty and is assumed to be Lipschitz continuous in $x(t) \in \mathcal{D}_x$. $ \mathcal{D}_x \subset \mathbb{R}^n$ is assumed to be a compact set and the control $u(t)$ is assumed to belong to a set of admissible control inputs of measurable and bounded functions \cite{ioannou1988theory,tao2003adaptive}. %, ensuring the existence and uniqueness of the solution to \eqref{eq:0}.

The flight control task is to have the closed-loop system follow a reference trajectory. There are many ways to generate the reference trajectory, including parameterizing it with polynomials or other smooth functions. Here we generate the reference trajectory using a linear reference model. This is a preferred approach in flight control because the desired transient and steady-state performance can be defined by a selecting the reference model's eigenvalues in the negative half plane \cite{calise:jgcd:00}. The desired closed-loop response of the reference system is given by
\begin{equation}
\dot x_{rm}(t) = A_{rm}x_{rm}(t) + B_{rm}r(t).
\label{eq:ref model}
\end{equation}
where $x_{rm}(t) \in \mathcal{D}_x  \subset \mathbb{R}^{n}$ and $A_{rm} \in \mathbb{R}^{n \times n}$ is Hurwitz and $B_{rm} \in \mathbb{R}^{n \times r}$. Furthermore, the command $r(t) \in \mathbb{R}^{r}$ denotes a bounded, piece wise continuous, reference signal and we assume the reference model (\ref{eq:ref model}) is bounded input-bounded output (BIBO) stable \cite{ioannou1988theory}.

The uncertainty function $\Delta(x)$ is unknown, but it is assumed to be piece-wise-continuous over a compact domain $\mathcal{D}_x \subset \mathbb{R}^n$, which is a reasonable assumption for flight control problems \cite{calise:jgcd:00,chowdhary2013guidance,Chowdhary:IJC:13}. We use a Deep Neural Networks (DNN) model to learn online a representation for this unknown function.  Using DNNs, a non linearly parameterized network estimate of the uncertainty can be written as $f_\theta(x) \triangleq \theta_n^T\Phi(x)$,  
% \begin{equation}
% \hat \Delta(x) \triangleq \hat W^T\phi(x)
% \label{eq:nn_estimate_defn}
% \end{equation}
where $\theta_n \in \mathbb{R}^{k \times m}$ are network parameters for the final layer and $\Phi(x)=\phi_n(\theta_{n-1},\phi_{n-1}(\theta_{n-2},\phi_{n-2}(...))))$, is a $k$ dimensional feature vector which is function of inner layer weights, activations and inputs. The basis vector $\Phi(x) \in \mathcal{F}: \mathbb{R}^{n} \to \mathbb{R}^{k}$ is considered to be Lipschitz continuous to ensure the existence and uniqueness of the solution (\ref{eq:0}).
%\gXX{is this description supposed to also include Deep networks? I think it would be a good idea to just mention how the basis can either be radial basis or created via a conjuction of bases as in deep networks, but here we focus on single layer networks to facilitate analysis}
\subsection{Adaptive Controller}
\label{adaptive_identification}
The aim of the adaptive controller is to construct a feedback law $u(t)$ such that the state of the uncertain dynamical system (\ref{eq:0}) asymptotically tracks the state of the reference model (\ref{eq:ref model}) in-spite of the unknown function $\Delta(x)$. In our formulation, an adaptive controller is designed to augment an existing baseline linear controller. Hence, the total control action consists of a linear feedback term $u_{pd} = Kx(t)$, a linear feed-forward term $u_{crm} = K_rr(t)$ and an adaptive term $\nu_{ad}(t)$: 
%\vspace{-2mm}
\begin{equation}
u = u_{pd} + u_{crm} - \nu_{ad}.
\label{eq:total_Controller}
\end{equation}
The baseline full state feedback and feed-forward controller (without the adaptive term $\nu_{ad}$)  is designed to ensure $A_{rm} = A-BK$ and $B_{rm} = BK_r$ when $\Delta(x)=0$. The intuition being that if the adaptive controller can ensure that $\nu_{ad}(t) \approx \Delta(x(t))$, then the closed loop system will essentially perform like the desired reference model. MRAC literature \cite{kannan:jgcd:05,Chowdhary2014IJC,gregory2009l1,chowdhary2013guidance} has relied on shallow networks to model the adaptive term $\nu_{ad}$. Here we present a framework to implement the DMRAC using DNN for a quadrotor control under unknown system faults using embedded computers.  %uncertainty information, we use a DNN estimate of the system uncertainties in the controller as $\nu_{ad}(t) = \hat{\Delta}(x(t))$.

\section{Deep Model Reference Adaptive Control} 
Unlike traditional MRAC or even single-hidden-layer neural network based MRAC weight update rule \cite{Chowdhary:JGCD:10,Lewis:AJC:99,kannan2005adaptive}, where the network weights are learned in the direction of minimizing the tracking error, training a deep Neural network is much more involved.  Feed-Forward networks like DNNs are trained in a supervised manner over a batch of i.i.d data. Deep learning optimization is based on Stochastic Gradient Descent (SGD) or its variants. 

The SGD update rule relies on a stochastic approximation of the expected value of the gradient of the loss function over a training set or mini-batches.
To train a deep network to estimate the system uncertainties, a batch of i.i.d samples of labeled pairs of state-true uncertainties $\{x(t),\Delta(x(t))\}$ are required. Since we do not have access to true values of $\Delta(x)$, we use the idea of using Model Reference Generative Network (MRGeN), introduced in \cite{joshi2018adaptive} to create estimates of $\Delta(x)$. This generative network is derived from separating the DMRAC-DNN into inner feature layer and the final output layer of the network. The parameter update of these two layers are temporally separated making it possible to online update the DMRAC, details of which are presented in following sections.
\subsection{DMRAC-Online Parameter Estimation law}
\label{Identification}
The last layer of DMRAC-DNN with learned features from inner layer forms the Deep-Model Reference Generative Network (D-MRGeN).  We use the classical MRAC learning rule to update pointwise in time the weights of the outermost layer of the DNN in the direction of achieving asymptotic tracking of the reference model by the actual system. This is now explained further in detail.

% Since we use the D-MRGeN estimates to train DNN model, we first study the admissibility and stability characteristics of the generative model estimate ${\Delta}'(x)$ in the controller (\ref{eq:total_Controller}).
%To achieve the asymptotic convergence of the reference model tracking error to zero, we use the DMRAC 
The adaptive term in the controller \eqref{eq:total_Controller} with a DNN model can be written as: %as $\nu_{ad} = \Delta'(x)$
\begin{equation}
    \nu_{ad}(t) = W^T\phi_n(\theta_{n-1},\phi_{n-1}(\theta_{n-2},\phi_{n-2}(...)))).
    \label{eq:adaptive_term}
\end{equation}
To differentiate the parameters of DMRAC from last layer weights ``$\theta_n$" of DNN,  we denote DMRAC weights as ``$W$". Recall, the goal of the MRAC controller is to ensure $\nu_{ad} \to \Delta(x)$ as $t \to \infty$.

\textbf{\emph{Assumption 1:}} Appealing to the universal approximation property of Neural Networks \cite{park1991universal} 
we have that, for every given basis functions $\Phi(x) \in \mathcal{F}$ there exists (unknown) ideal weights $W^* \in \mathbb{R}^{k \times m}$ and $\epsilon_1(x) \in \mathbb{R}^{m}$ such that the following
approximation holds
\begin{equation}
\Delta(x) = W^{*T}\Phi(x) + \epsilon_1(x), \hspace{2mm} \forall x(t) \in \mathcal{D}_x \subset \mathbb{R}^{n}
\label{eq:3}
\end{equation}

The network approximation error $\epsilon_1(x)$ is upper bounded, s.t  $\bar{\epsilon}_1 = \sup_{x \in \mathcal{D}_x}\|\epsilon_1(x)\|$, and can be made arbitrarily small given sufficiently large number of basis functions \cite{cybenko1989approximation}

We can define the reference model tracking error as $e(t) = x_{rm}(t)- x(t)$.
Using (\ref{eq:0}) \& (\ref{eq:ref model}) and the controller of form (\ref{eq:total_Controller}) with adaptation term $\nu_{ad}$ (\ref{eq:adaptive_term}), the tracking error dynamics can be written as  
\begin{equation}
\dot e(t) = A_{rm}e(t) + \tilde W^T\Phi(x) +  \epsilon_1(x).
\label{eq:14}
\end{equation}
where $\tilde W = W^*-W$ is error in parameter. The estimate of the unknown true network parameters $W^*$ are calculated on-line using the weight update rule (\ref{eq:18}); correcting the weight estimates in the direction of minimizing the instantaneous tracking error $e(t)$.  The resulting update rule for network weights in estimating the total uncertainty in the system is as follows \cite{Ioannou:96bk,ioannou1988theory,tao2003adaptive}, 
\begin{equation}
\dot {W} = \Gamma proj(W,\Phi(x)e(t)'P), \hspace{5mm} {W}(0) = {W}_0 \label{eq:18}
\end{equation} 
where $proj$ is a projection operator, $\Gamma \in \mathbb{R}^{k \times k}$ is the learning rate and $P \in \mathbb{R}^{n \times n}$ is a positive definite matrix such that for a given Hurwitz $A_{rm}$, the matrix $P \in \mathbb{R}^{n \times n}$ is a positive definite solution of Lyapunov equation $A_{rm}^TP + PA_{rm} + Q = 0$ for given $Q > 0$.
% \begin{equation}
% A_{rm}^TP + PA_{rm} + Q = 0
% \end{equation}

\textbf{\emph{Assumption 2:}} For uncertainty parameterized by unknown true weight ${W}^* \in \mathbb{R}^{k \times m}$ and known nonlinear basis $\Phi(x)$,
the ideal weight matrix is assumed to be upper bounded s.t $\|{W}^*\| \leq \mathcal{W}_b$. 
% This is not a restrictive assumption. 

% \subsubsection{Lyapunov Analysis}
% The on-line adaptive identification law (\ref{eq:18}) guarantees the asymptotic convergence of the tracking errors $e(t)$ and parameter error $\tilde W(t)$ under the condition of persistency of excitation \cite{aastrom2013adaptive,ioannou1988theory} for the structured uncertainty. Similar to the results by Lewis for SHL networks \cite{lewis1999nonlinear}, we show here that under the assumption of unstructured uncertainty represented by a deep neural network, the tracking error is uniformly ultimately bounded (UUB). We will prove the following theorem under switching feature vector assumption.
% Under this problem formulation and assumptions, the proof of the following theorem is given in \cite{joshi2019deep}:
\begin{theorem}
    Consider the actual and reference plant model (\ref{eq:0}) \& (\ref{eq:ref model}). If the weights parameterizing total uncertainty in the system are updated according to identification law \eqref{eq:18} Then the tracking error $\|e\|$ and error in network weights $\|\tilde W\|$ are bounded for all $\Phi \in \mathcal{F}$.
\label{Theorem-1}
\end{theorem} 
The proof of above theorem is provided in the Appendix-\ref{Appendix:proof_of_theorem_1}.
From Theorem-\ref{Theorem-1} \& (\ref{eq:14}) and using system theory \cite{kailath1980linear} we can infer that as $e(t) \to 0$, and $\nu_{ad} \to \Delta(x)$ in point-wise sense. Hence uncertainty estimates $y_{\tau} = \nu_{ad}(x_{\tau})$ are admissible target values for training DNN features over the data $Z^M = \{\{x_{\tau},y_{\tau}\}\}_{\tau = 1}^{M}$. 

The details of DNN training and implementation details of DMRAC controller is presented in the following section.

\subsection{Deep Feature Training for DMRAC controller}
This section provides the details of the DNN training over data samples observed over n-dimensional input subspace $x(t) \in \mathcal{X} \in \mathbb{R}^{n}$ and m-dimensional target subspace $y\in\mathcal{Y} \in \mathbb{R}^m$. The sample set is denoted as $\mathcal{Z}$ where $\mathcal{Z} \in \mathcal{X} \times \mathcal{Y}$.

We are interested in the function approximation task for DNN. Let the function $f_{\boldsymbol{\theta}}(x)= \theta_n\phi_n(\theta_{n-1},\phi_{n-1}(\theta_{n-2},\phi_{n-2}(...))))$ s.t $f_{\boldsymbol{\theta}}: \mathbb{R}^n \to \mathbb{R}^m$ be the network approximating the model uncertainty with parameters $\boldsymbol{\theta} \in \boldsymbol{\Theta}$, where $\boldsymbol{\Theta}$ is the space of parameters. We assume a training data buffer $\mathcal{B}$ can store $p_{max}$ training examples, such that the set $Z^{p_{max}} = \{Z_i | Z_i \in \mathcal{Z}\}_{i=1}^{p_{max}} = \{(x_i, y_i) \in \mathcal{X}\times\mathcal{Y}\}_{i=1}^{p_{max}}$. A batch of samples can be drawn independently from the buffer $\mathcal{B}$ over probability distribution $P$ for DNN training.
% The hypothesis set, which consist of all possible functions $f_{\boldsymbol{\theta}}$ is denoted as $\mathcal{H}$. Therefore a learning algorithm $\mathcal{A}$ (in our case SGD) is a mapping from $\mathcal{A}: \mathcal{Z}^{p_{max}} \to \mathcal{H}$ 

The loss function, which measures the discrepancy between true target $y$ and algorithm's estimated target function value $f_{\boldsymbol{\theta}}$ is denoted by $L(y, f_{\boldsymbol{\theta}}(x))$. Specific to work presented in this paper, we use a $\ell_2$-norm between values i.e. $\mathbb{E}_p(\ell(y, f_{\boldsymbol{\theta}}(x))) = \mathbb{E}_P \left(\|y_i - f_{\boldsymbol{\theta}}(x_i)\|_2\right)$ as loss function for DNN training. An empirical loss $L(\boldsymbol{Z, \theta}) = \frac{1}{M}\sum_i^M \ell(\boldsymbol{Z_i, \theta})$ is used to approximate the loss function since the true distribution $P$ is unknown to learning algorithm. The network parameters are updated using SGD in the direction of minimizing the loss function as follows  \begin{eqnarray}
 \boldsymbol{\theta}_{k+1} &=& \boldsymbol{\theta}_k - \eta
 \frac{1}{M}\sum_i^M \nabla_{ \boldsymbol{\theta}}L(\boldsymbol{\theta}).
 \label{SGD2}
 \end{eqnarray}
Unlike the conventional DNN training where the true target values $y \in  \mathcal{Y}$ are available for every input $x \in  \mathcal{X}$, in DMRAC true system uncertainties as the labeled targets are not available for the network training. We use the part of the network itself (the last layer) with pointwise weight updated according to MRAC-rule as the generative model for the data. The DMRAC uncertainty estimates $y = {W}^T\Phi(x,\theta_1,\theta_2, \ldots \theta_{n-1})$ along with inputs $x_i$ make the training data set $Z^{p_{max}} = \{x_i, {\Delta}'(x_i)\}_{i=1}^{p_{max}}$. Note that we use interchangably $x_i$ and $x(t)$ as discrete representation of continuous state vector for DNN training. The main purpose of DNN in the adaptive network is to extract relevant features of the system uncertainties, which otherwise is very tedious to obtain with unknown bounds of the domain of operation.

We also analyse the features through data visualization techniques \cite{maaten2008visualizing} and demonstrate empirically, that the DNN features trained over past i.i.d representative data retains the memory of the past instances and can be used as the frozen feed-forward network over similar reference tracking tasks without loss of the guaranteed tracking performance.

\subsection{Recording Data using MRGeN for DNN Training}
\label{subsection-buffer}
The DNN in DMRAC controller is trained over training dataset $Z^M = \{x_i, y_i\}_{i=1}^M$, where the $y_i$ are fast updating outer layer of DNN estimates of the uncertainty. The training dataset $Z^M$ is randomly drawn from a larger data buffer $\mathcal{B}$. Not every pair of data $\{x_i, y_i\}$ from generative network is added to the training buffer $\mathcal{B}$. We qualify the input-target pair based on kernel independence test to ensure that we collect locally exciting independent information which provides a sufficiently rich representation of the operating domain. Since the state-uncertainty data is the realization of a Markov process, such a method for qualifying data to be sufficiently independent of previous data-points is necessary. The algorithm details to qualify and add a data point to the buffer is provided in detail in \cite{5991481}.
\begin{algorithm}[h!]
    \caption{D-MRAC Controller Training}
    \label{alg:DMRAC}
    \begin{algorithmic}[1]
        \STATE {\bfseries Input:} $\Gamma, \eta, \zeta_{tol}, p_{max}$
        \WHILE{New measurements are available}
        \STATE Update the DMRAC weights $W$ using Eq:\eqref{eq:18}
        \STATE Compute $y_{\tau+1} = \hat{W}^T\Phi(x_{\tau+1})$
        \STATE Given $x_{\tau+1}$ compute $\gamma_{\tau+1}$ \cite{5991481}.
                \IF{$\gamma_{\tau+1} \geqslant \zeta_{tol}$}
                \STATE Update $\mathcal{B}:\boldsymbol{Z}(:) = \{x_{\tau+1}, y_{\tau+1}\}$ and $\mathbb{X}: \Phi(x_{\tau+1})$
                	\IF{$|\mathcal{B}| > p_{max}$}
                    \STATE Delete element in $\mathcal{B}$ by SVD maximization \cite{5991481}
                    \ENDIF
        	    \ENDIF
    	    \IF{$|\mathcal{B}| \geq M$}
    	    \STATE Sample a mini-batch of data $\boldsymbol{Z}^M \subset \mathcal{B}$
    	    \STATE Train the DNN network over mini-batch data using Eq-\eqref{SGD2}
    	    \STATE Update the feature vector $\Phi$ for D-MRGeN network
    	    \ENDIF
        \ENDWHILE
        \end{algorithmic}
\end{algorithm}

\section{DMRAC Flight control Experimental Setup}
\begin{figure*}[h]
    \centering
    \includegraphics[width = 0.8\textwidth]{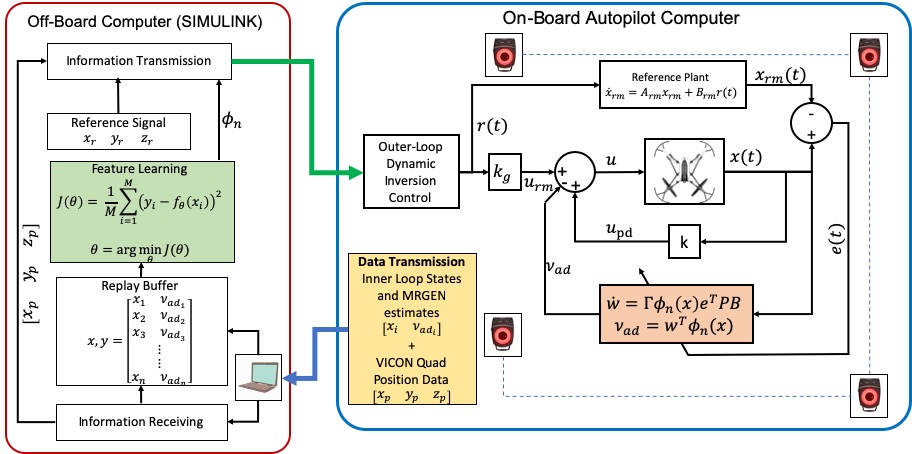}
    \caption{Our On-board - Off-board Implementation of Deep Model reference Adaptive controller for Quadrotor control. The system ensures that the outer layer weights can be updated onboard in a manner to ensure Lyapunov stability (at 200Hz), while the inner layer weights are updated asynchronously off-board to improve learning performance.}
    \label{fig:onboard-offboard_controller}
\end{figure*}

This section provides the hardware and vehicle details for evaluation of DMRAC for flight control of quadrotor system.
The flight tests were done on a commercially available quadcopter
platform, Parrot Mambo Mini-Drone in a VICON facility. 
This drone was chosen for flight testing due to its small size, low cost, and relative ease of implementing and testing algorithms. The implementation of DMRAC was done using our onboard, off-board architecture as shown in Figure-\ref{fig:onboard-offboard_controller}. The off-board component consists of software modules which run entirely on a host computer and are responsible for training the neural network using data received from the drone via UDP bluetooth protocol. The data received are the drone's current roll, pitch, yaw angles and the angular rates, and onboard estimate of the total uncertainty using the features provided from off-board computer. The above information is stored in a memory buffer and random batches from this buffer are used to train the deep neural network on the off-board computer. Moreover, the host computer also receives drone's current position inside the Vicon Arena from the Vicon system via Wi-Fi which is then communicated to
\noindent  the drone using UDP protocol. Also, as soon as the inner layer network parameters are updated using the data collected from the drone, the host computer communicates the updated inner-layer parameters back to the vehicle, which are used in updating the DMRAC controller to produce adaptive torque. The onboard software modules are responsible for state estimation along with computing the entire control effort. The main reason behind spitting the controller into on-board and off-board module is because of the 
limited processing power of the drone. It is possible that if a more powerful processor can be embedded on-board, the algorithm can be implemented completely on the vehicle. Even in this case however, it is desirable to separate the inner layer learning so that the outer layer weights can be updated to ensure Lyapunov stability and inner layer weight only updated whenever the test data loss values are significantly high. Furthermore, in this case, an interesting extension could be to use the presented on-board - off board architecture to asynchronously update the inner-layer weights using data from multiple drones simultaneously. However, this is left for future work.

\section{Fault tolerance: Rotor blade chipping in mid-flight}
\begin{figure}[tbh!]
     \centering
     \includegraphics[width=0.7\columnwidth]{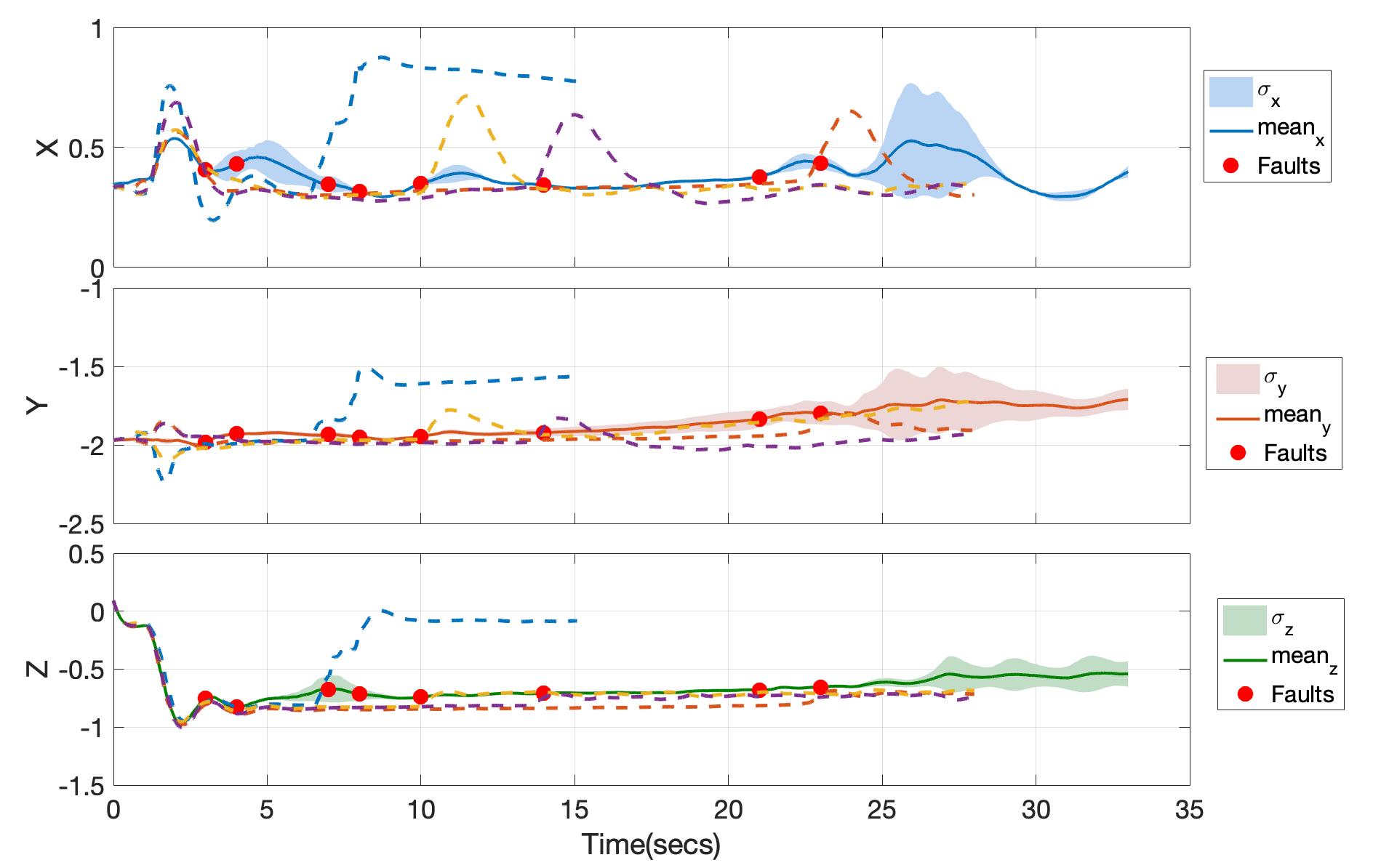}
     \caption{MRAC Trajectory tracking performance in X-Y-Z under system fault for eight flight test. Out of eight flights we observe four times the quadrotor either crashed or produced bad tracking}
     \label{fig:mrac_quad_chipped}
\end{figure}

\begin{figure}[tbh!]
     \centering
     \includegraphics[width=0.7\columnwidth]{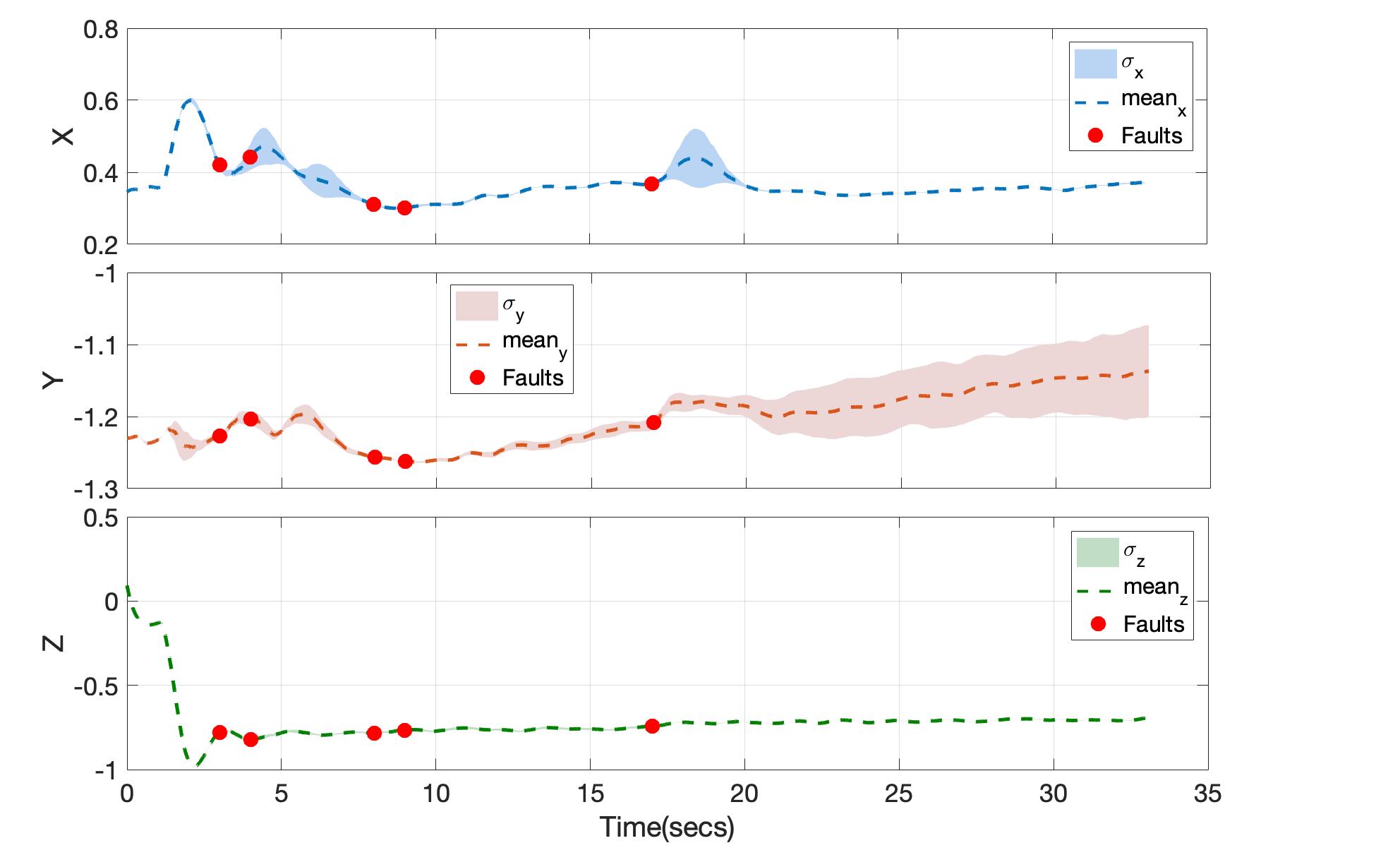}
     \caption{DMRAC Trajectory tracking performance in X-Y-Z under system fault for Eight flight test (Red dot: Time at which Fault occurred)}
     \label{fig:dmrac_quad_chipped}
\end{figure}

To test the fault-tolerance capability of the controllers, we test and compare the performance of PID, MRAC, and DMRAC when rotor chipping occurs during mid-flight. The quadrotor is commanded to hover at $1m$ above the ground. Due to centrifugal forces, the chipped blade breaks off and causes the fault into the system at an undertermined time. Since this is not a controlled fault, to ensure the reliability of the controller, we report controller performance over multiple runs. The results presented in Fig-\ref{fig:dmrac_quad_chipped} clearly shows that DMRAC outperforms PID and MRAC. In the case of PID, only two runs were carried out, since in both cases, the drone underwent severe oscillation and crashed. Tests conclusively demonstrated that PID is not capable of handling extreme faults in the system even with extensive tuning (We do not provide the tracking plots for PID but we include the flight video for all three controllers\footnote{\url{https://girishvjoshi.github.io/gj_blog/dmrac/update/2019/12/18/Deep-Model-Reference-Adaptive-Control.html}}). In the case of MRAC and DMRAC, eight flight tests are carried out. Out of eight test runs, MRAC managed to control only in four flights. Whereas no failure were observed in the case of DMRAC. Also, on comparing flights where no crash occurred, one can see that MRAC produces poor reference tracking when compared to DMRAC. The Figure-\ref{fig:mrac_quad_chipped} and \ref{fig:dmrac_quad_chipped} show for successful flights (8 flights for DMRAC and 4 for MRAC) mean and variance of the reference tracking in the x-y-z position for each algorithm. The dotted lines in MRAC plots are the state history for failed flights. Additional results with high wind bias under a nonlinear disturbance, sim-to-real transfer and learning retention experiments are provided in Appendix-\ref{Appendix:additional_results}.
\section{Feature Analysis of DMRAC controller}
Analyzing a deep network success in performing inference tasks is mathematically not feasible and has become a challenging question. We often treat the deep neural network models as black boxes without clear understanding on internal mechanics. Understanding how the network is able to store and present adaptive controller the appropriate features for the appropriate flight regime is a hard question. As a result, we face a challenge that demand visualizing these high dimension network data for better understanding and analyzing machine learning models, especially their inner working mechanism. 

Principle Component Analysis(PCA) \cite{wold1987principal} and t-Distributed Stochastic Neighborhood Embedding (t-SNE) \cite{maaten2008visualizing} are two very important techniques in the point based visualization of higher dimensional data in to lower dimensional feature vectors. These methods will help us explore and analyze the relationship between the cluster in the deep feature for different flight regime the network has seen through training. 

We conducted the flight test with varying wind bias cases and rotor breaking case (videos in supplementary). We deployed a DMRAC network to control a physical quadrotor tracking figure-8 reference signal under wind bias and Rotor fault. The same DNN is trained progressively over the ensemble of data sampled from Low, Medium, High wind bias and Fault tolerant flight cases.
\begin{figure*}[tbh!]
	\centering
	\begin{subfigure}{0.46\columnwidth}
		\includegraphics[width=\textwidth]{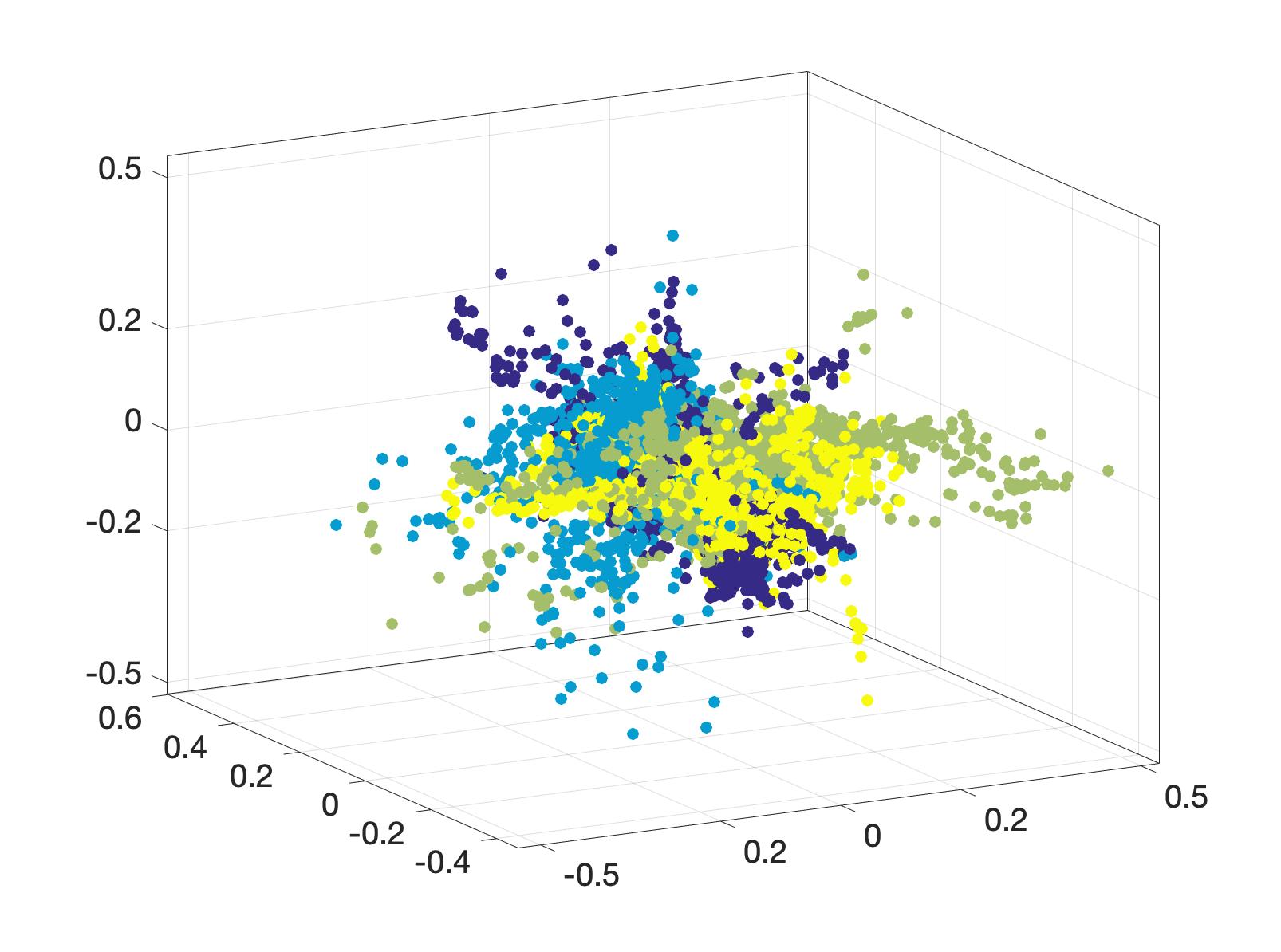}
		\caption{}
		\label{fig:Quad_trajectory}
	\end{subfigure}
	\begin{subfigure}{0.46\columnwidth}
		%\centering
		\includegraphics[width=\textwidth]{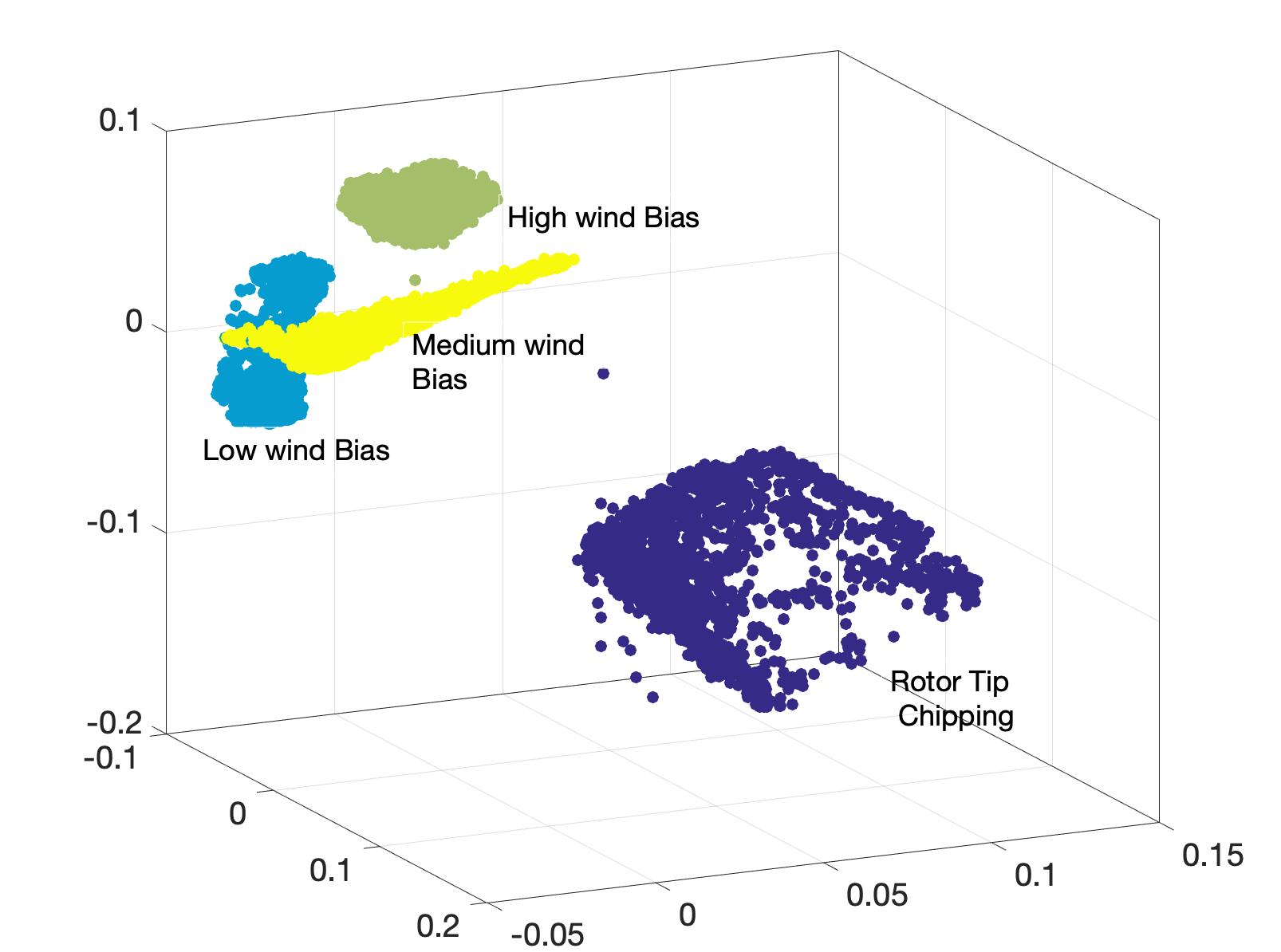}
		%\vspace{-0.2in}
		\caption{}
		\label{fig:chap8_quad_tnse}
	\end{subfigure}
	\caption{t-SNE visualization of DMRAC features (a) Before training and (b) After training to handle Low, Medium, High Wind Bias and Rotor Tip breaking case}
	\label{fig:feature_visualization}
\end{figure*}

Figure-\ref{fig:chap8_quad_tnse} show the t-SNE visualization of the $20-$dimensional feature vector in $3-$dimensional scatter-plot. In Figure-\ref{fig:chap8_quad_tnse} each point denote one feature representation of test state from either of the experiments. As shown in the Figure-\ref{fig:feature_visualization}, after training, the spatial separation between classes is significantly improved. This observation provides evidence to our hypothesis that neural networks learn to detect representations that are useful for different flight regimes and uncertainties experienced by the vehicle. The feature visualization also helps with the understanding of similar flight regimes. As we can observe the windy and Fault cases are classified as two very distinct clusters. But the wind cases from Low to Medium to High wind case cluster are together but progressively move away from each other as the wind disturbance is increased.

\section{Conclusion}
\label{conclusions}
In this paper, we presented a system to implement DMRAC adaptive controllers using model reference generative network architecture. % to address the issue of feature design in unstructured uncertainty. 
We demonstrated that our fast-slow architecture utilizing asynchronous onboard and off-board processing can be used to incorporate deep learning in the closed-loop for high-bandwidth flight control of unstable aircraft in the presence of significant disturbances. The results clearly show that when utilized in the closed loop, DMRAC can provide significant performance and generalization benefits over shallow MRAC and PIDs. The results are significant, not only for flight control, but for other robotic control applications involving deep learning. This is because our approach of separating the learning in asynchronous manner can be adopted to other learning based controllers, including learning based MPC and reinforcement learning.
%The proposed controller uses DNN to model significant uncertainties without knowledge of the system's domain of operation. We provide theoretical proofs of the controller generalizing capability over unseen data points and boundedness properties of the tracking error. Numerical simulations with 6-DOF quadrotor model demonstrate the controller performance, in achieving reference model tracking in the presence of significant matched uncertainties and also learning retention when used as a feed-forward adaptive network on similar but unseen new tasks. Thereby we claim DMRAC is a highly powerful architecture for high-performance control of nonlinear systems with robustness and long-term learning properties.

% \section*{Acknowledgments}
%% Use plainnat to work nicely with natbib.
\bibliographystyle{unsrt}
\bibliography{references_gc,daslab_all,daslab_pubs,references,references_gj}
\newpage
\appendix
\section{Appendices}
\subsection{Network Partitioning details for On-board and Off-board training.}
\begin{figure}[tbh!]
    \centering
	\begin{subfigure}{0.485\columnwidth}
		\includegraphics[width=1\columnwidth]{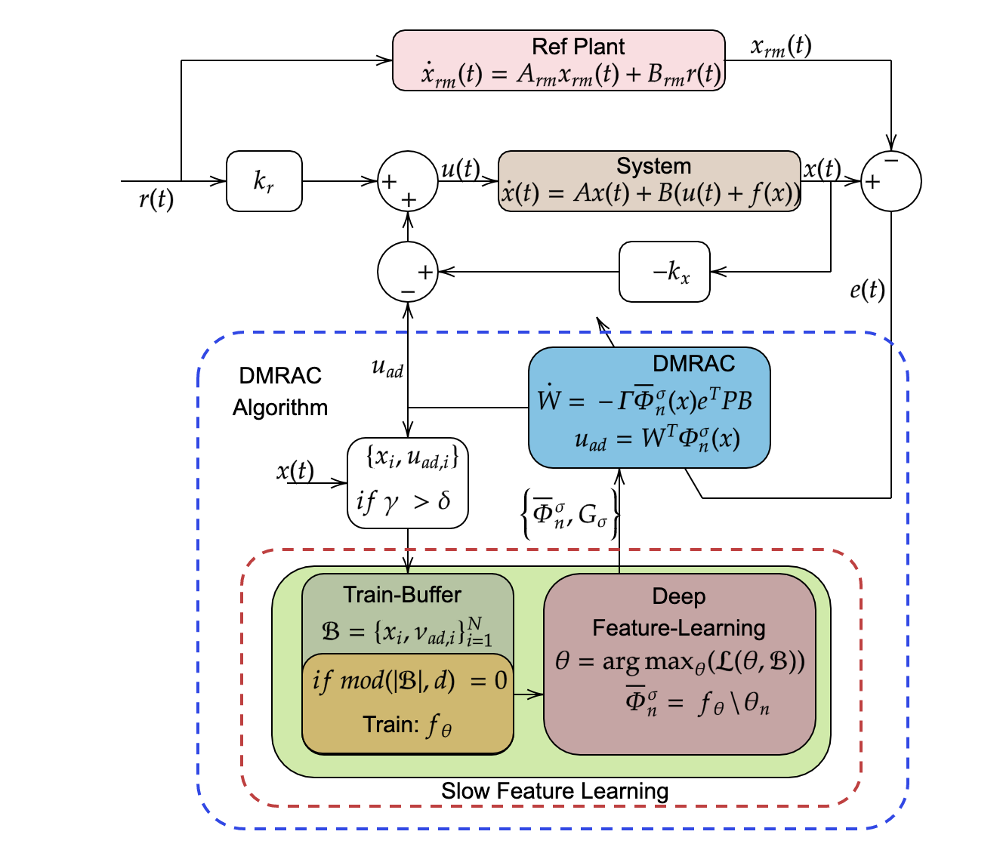}
		\caption{}
		\label{fig:DNN_architecture}
	\end{subfigure}
	\begin{subfigure}{0.485\columnwidth}
		\includegraphics[width=1\columnwidth]{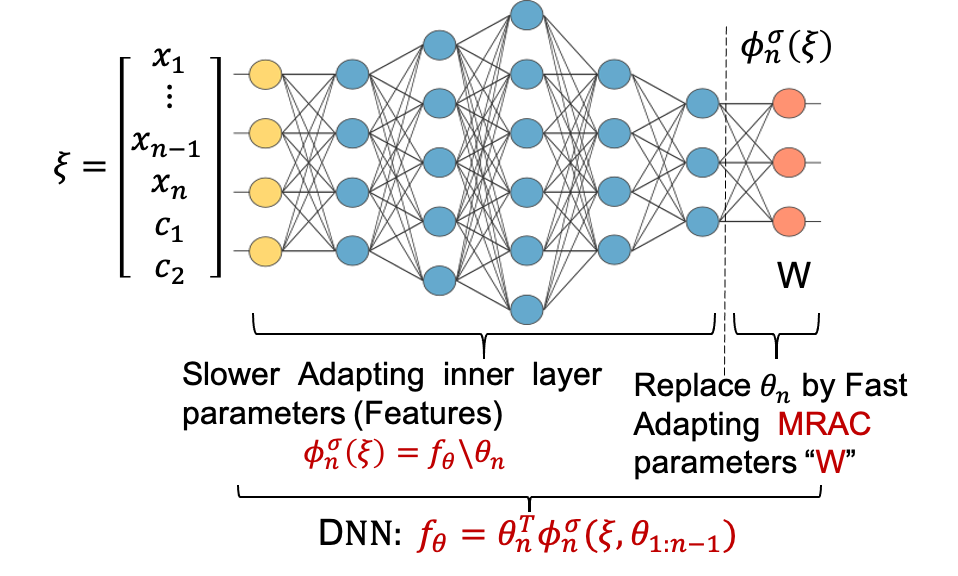}
		\caption{}
		\label{fig:DNN_network}
	\end{subfigure}
	\vspace{-0.12in}
	\caption{DMRAC Controller Architecture (a) Over-all Asynchronous DMRAC training block diagram (b) Network Partitioning scheme for On-board and Off-board update.}
	\label{fig:grid_world}
\end{figure}

The DNN architecture for MRAC is trained in two steps. We separate the DNN into two networks as shown in Fig-\ref{fig:DNN_architecture}. The faster learning outer adaptive network and slower deep feature network. DMRAC learns underlying deep feature vector to the system uncertainty using locally exciting uncertainty estimates obtained using a generative network. Between successive updates of the inner layer weights, the feature provided by the inner deep network is used as the fixed feature vector for outer layer adaptive network update and evaluation Figure-\ref{fig:DNN_network}. The algorithm for DNN learning and DMRAC controller is provided in Algorithm-\ref{alg:DMRAC}. Through this architecture of mixing two-time scale learning, we fuse the benefits of DNN memory through retention of relevant, exciting features and robustness, boundedness guarantee in reference tracking. This key feature of the presented framework ensures robustness while guaranteeing long term learning and memory in the adaptive network.

Also as indicated in the controller architecture Fig-\ref{fig:DNN_architecture} we can use contextual state `$c_i$' other than system state $x(t)$ to extract relevant features. These contextual states could be relevant model information not captured in system states. For example for an aircraft system, vehicle parameters like pitot tube measurement, the angle of attack, engine thrust and so on. These contextual states can extract features which help in decision making in case of faults. The work on DMRAC with contextual states will be dealt with in the follow on work.

\subsection{Proof of Theorem-1}
\label{Appendix:proof_of_theorem_1}
\textbf{\textit{Theorem-1}} Consider the actual and reference plant model (1) \& (2). If the weights parameterizing total uncertainty in the system are updated according to identification law (3) Then the tracking error $\|e\|$ and error in network weights $\|\tilde W\|$ are bounded for all $\Phi \in \mathcal{F}$.

\begin{proof}
The feature vectors belong to a function class characterized by the inner layer network weights $\theta_i$ s.t $\Phi \in \mathcal{F}$. We will prove the Lyapunov stability under the assumption that inner layer of DNN presents us a feature which results in the worst possible approximation error compared to network with features before switch. 

For the purpose of this proof let $\Phi(x)$ denote feature before switch and $\bar{\Phi}(x)$ be the feature after switch. We define the error $\epsilon_2(x)$ as,
\begin{equation}
    \epsilon_2(x) = \sup_{\bar{\Phi} \in \mathcal{F}}\left|W^T\bar{\Phi}(x) - W^T\Phi(x)\right|
\end{equation}
Similar to \textbf{\emph{Fact-1}} we can upper bound the error $\epsilon_2(x)$ as $\bar{\epsilon}_2 = \sup_{x \in \mathcal{D}_x}\|\epsilon_2(x)\|$.
By adding and subtracting the term $W^T\bar{\Phi}(x)$, we can rewrite the error dynamics (6) with switched basis as,
\begin{eqnarray}
\dot e(t) &=& A_{rm}e(t) + W^{*T}\Phi(x) - W^T\Phi(x) \nonumber \\
&& + W^T\bar{\Phi}(x) - W^T\bar{\Phi}(x) + \epsilon_1(x)
\label{eq:14_1}
\end{eqnarray}
From \textbf{\emph{Assumption-1}} we know there exists a $W^*$ $\forall \Phi \in \mathcal{F}$. Therefore we can replace $W^{*T}\Phi(x)$ by $W^{*T}\bar{\Phi}(x)$ and rewrite the Eq-\eqref{eq:14_1} as
\begin{eqnarray}
\dot e(t) &=& A_{rm}e(t) + \tilde{W}^{T}\bar{\Phi}(x) + W^T(\bar{\Phi}(x) - \Phi(x)) + \epsilon_1(x) \nonumber \\
\label{eq:14_2}
\end{eqnarray}
For arbitrary switching, for any $\bar{\Phi}(x) \in \mathcal{F}$, we can prove the boundedness by considering worst possible approximation error and therefore can write,
\begin{eqnarray}
\dot e(t) &=& A_{rm}e(t) + \tilde{W}^{T}\bar{\Phi}(x) +  \epsilon_2(x) + \epsilon_1(x) 
\label{eq:14_3}
\end{eqnarray}
Now lets consider $V(e,\tilde W) > 0$ be a differentiable, positive definite radially unbounded Lyapunov candidate function,
%ensuring the asymptotic convergence of tracking error $e(t)$ and error in network weight $\tilde W$
\begin{equation}
V(e,\tilde W) = e^TPe + \frac{\tilde W^T \Gamma^{-1} \tilde W}{2}
\label{eq:20}
\end{equation}
%where $\Gamma >0$ is the adaption rate.
The time derivative of the lyapunov function (\ref{eq:20}) along the trajectory (\ref{eq:14_3}) can be evaluated as
\begin{equation}
\dot V(e,\tilde W) = \dot e^TPe + e^TP \dot e - \tilde W^T\Gamma^{-1}\dot{\hat W}
\label{eq:25}
\end{equation}
% \begin{eqnarray}
% \dot V(e,\tilde W) &=& -e^TQe + \left(\phi(x)e'P - \Gamma^{-1}\dot{\hat W}\right)2\tilde W \nonumber\\
% && + 2e^TP\epsilon(x) 
% \label{eq:21}
% \end{eqnarray}
Using (\ref{eq:14_3}) \& (\ref{eq:18}) in (\ref{eq:25}), the time derivative of the lyanpunov function reduces to 
\begin{eqnarray}
\dot V(e,\tilde W) &=& -e^TQe + 2e^TP\epsilon(x)
\label{eq:22}
\end{eqnarray}
where $\epsilon(x) = \epsilon_1(x) +\epsilon_2(x)$ and $\bar{\epsilon} = \bar{\epsilon_1} + \bar{\epsilon_2}$.\\ 
Hence $\dot V(e,\tilde W) \leq 0$ outside compact neighborhood of the origin $e = 0$, for some sufficiently large $\lambda_{min}(Q)$. 
\begin{equation}
\|e(t)\| \geq \frac{2\lambda_{max}(P)\bar\epsilon}{\lambda_{min}(Q)}
\label{eq:error_bound}
\end{equation}
Using the BIBO assumption $x_{rm}(t)$ is bounded for bounded reference signal $r(t)$, thereby $x(t)$ remains bounded.  Since $V(e,\tilde W)$  is radially unbounded the result holds for all $x(0) \in \mathcal{D}_x$.
Using the fact, the error in parameters $\tilde{W}$ are bounded through projection operator \cite{larchev2010projection} and further using Lyapunov theory and Barbalat’s Lemma \cite{narendra2012stable} we can show that $e(t)$ is uniformly ultimately bounded in vicinity to zero solution. 
\end{proof}

\subsection{Additional Results}
\label{Appendix:additional_results}
In this section, we present flight test results that compare the performance of DMRAC algorithm over
other popular flight control algorithms like MRAC and PID. 

\subsubsection{Trajectory tracking under highly non linear wind disturbance}
\begin{figure*}[tbh]
    \centering
    \includegraphics[width=0.9\linewidth]{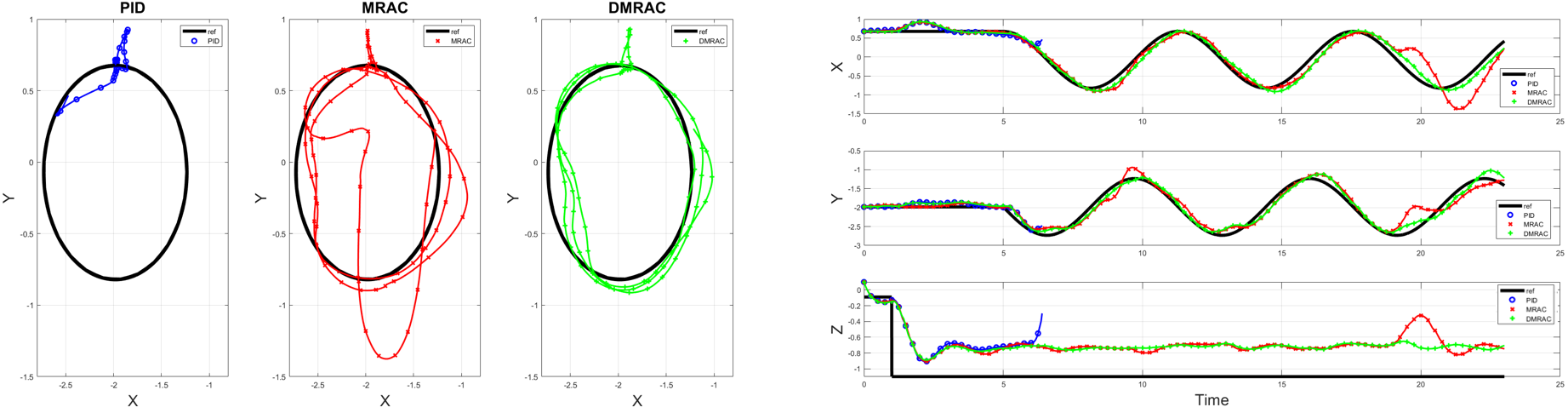}
    \caption{Tracking performance of quadrotor in x-y-z, for a circular trajectory under high wind bias with nonlinear and unpredictable disturbance.}
    \label{circle_wind_cloth_xy}
\end{figure*}

\begin{figure*}[!tbh]
    \centering
    \includegraphics[width=0.9\linewidth]{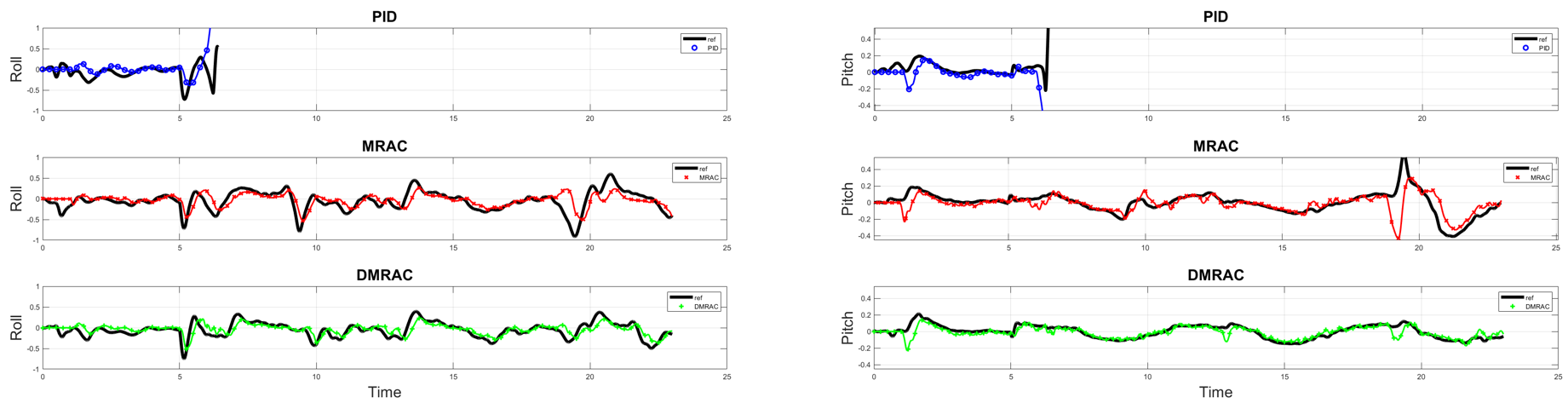}
    \caption{Tracking performance of quadrotor in Roll-Pitch, for a circular trajectory under wind bias with nonlinear and unpredictable disturbance.}
    \label{circle_wind_cloth_rpy}
\end{figure*}

\begin{figure*}[!tbh]
    \centering
    \includegraphics[width=0.9\linewidth]{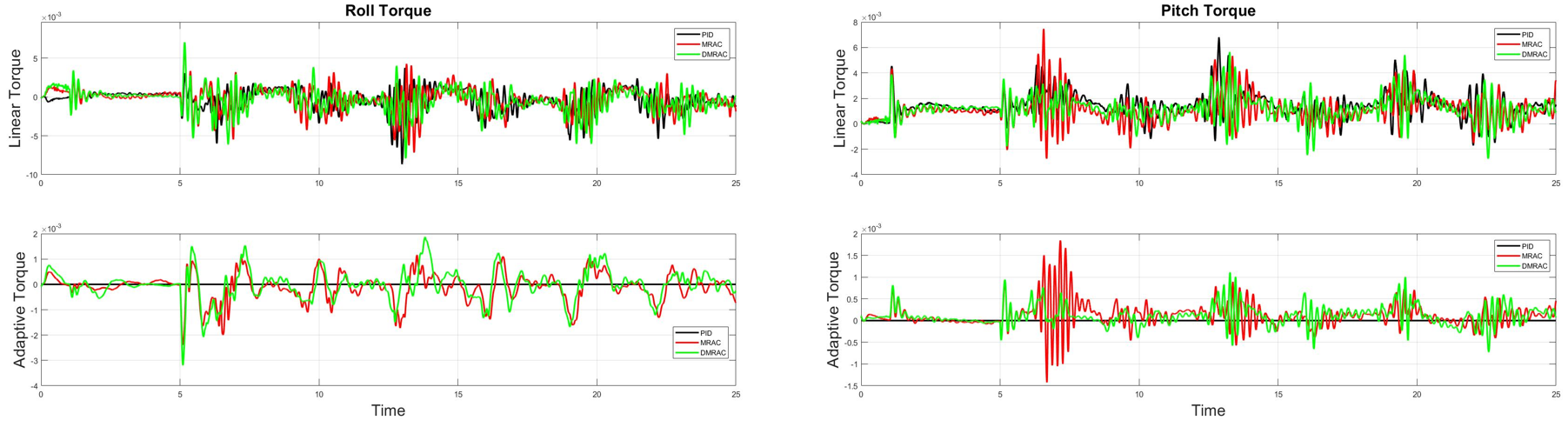}
    \caption{Feedback,Feed-forward and Adaptive control torque for PID, MRAC and DMRAC for Circular trajectory tracking under high wind bias}
    \label{circle_wind_cloth_torque}
\end{figure*}

In this experiment, we simulated a highly nonlinear and unpredictable external disturbance on the quadrotor. We attached a piece of cloth underneath the frame of the vehicle, which is then subjected to high wind bias. The erratic flapping of the cloth produces unpredictable disturbance torques and forces on the quadrotor. The experiment is designed to push each controller to their limits and is conducted thrice to demonstrate repeatability. Here, we present the best case performance of each controller. We observe that PID fails very early in the experiment, whereas both the adaptive controllers give a stable performance. However, the tracking error observed for (shallow) MRAC was relatively higher when compared to DMRAC. Refer Figure-\ref{circle_wind_cloth_xy} and Figure-\ref{circle_wind_cloth_rpy}. Figure-\ref{circle_wind_cloth_torque} shows plots for control torques generated in for each algorithm for the above case.

\subsubsection{Evaluating Transfer Learning with DMRAC}
\begin{figure}[!htb]
    \centering
    \includegraphics[width=0.75\columnwidth]{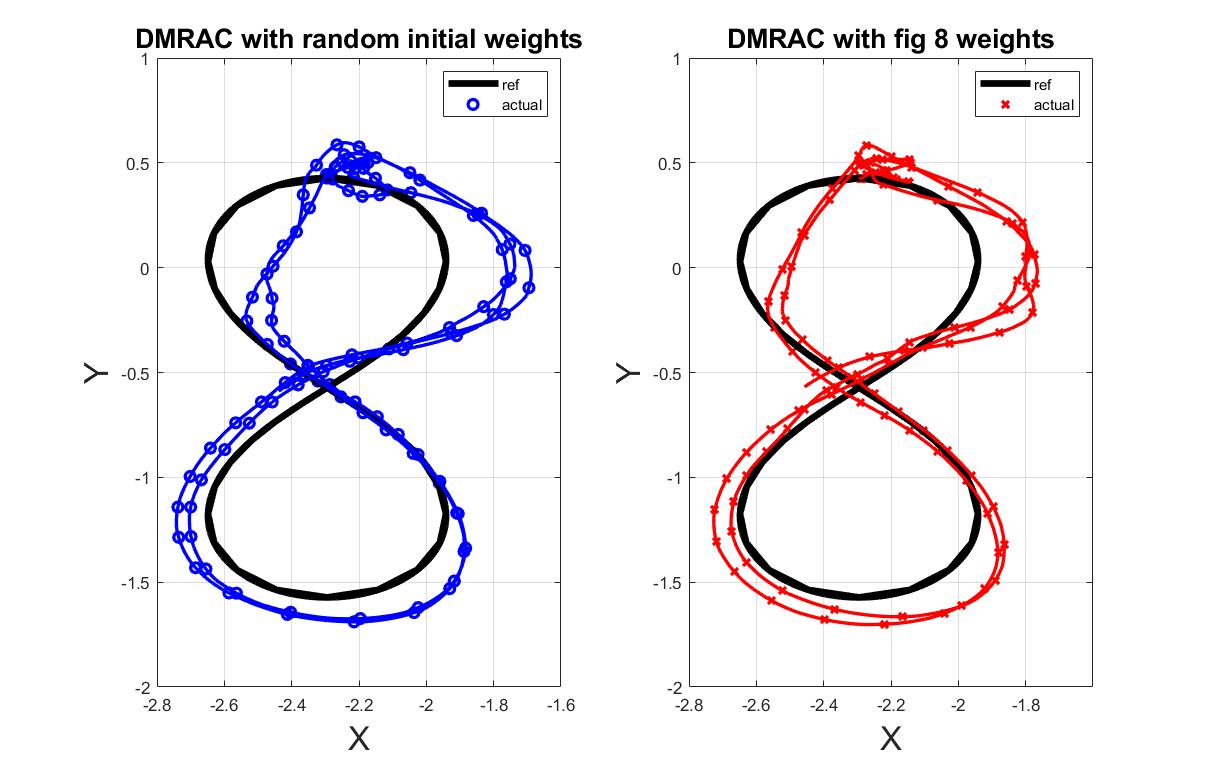}
    \caption{Figure of 8 Trajectory tracking under wind bias with random initialization vs. Feature transfer in DMRAC}
    \label{fig:DMRAC_TL_xy}
\end{figure}

Lately, Transfer learning (TL) has been a much-researched topic in machine learning and reinforcement learning. In similar lines in these experiments, we aim to test the advantages of representation transfer in an adaptive control setting. We test transfer learning through sharing the network parameters between tasks. TL is performed by first running DMRAC on related tasks and learning the network weights, which incorporate some feature knowledge. These learned weights are then used to initialize a fresh DMRAC network executing a new unseen task.  We use a flight test of a drone executing a basic figure of 8 trajectories as a source task for representation transfer through the warm-start of the networks. The target task is an unseen but related task for which an initialized network is used for the drone executing figure of 8 trajectories under high wind bias, refer Fig-\ref{fig:DMRAC_TL_xy}. A clear improvement of controller performance in achieving smaller transients and better steady-state tracking is observed with warm-started DMRAC. The deep network weights learnt over the quadrotor executing figure of 8 trajectories encodes the feature knowledge about modeling uncertainties. When we transfer this learning to a new drone executing figure of 8 with wind bias, it is able to adapt faster and also quickly learn features corresponding to wind bias.
\subsubsection{Simulation to Real-World Transfer Learning}
\begin{figure}[!tbh]
    \centering
    \includegraphics[width=0.75\columnwidth]{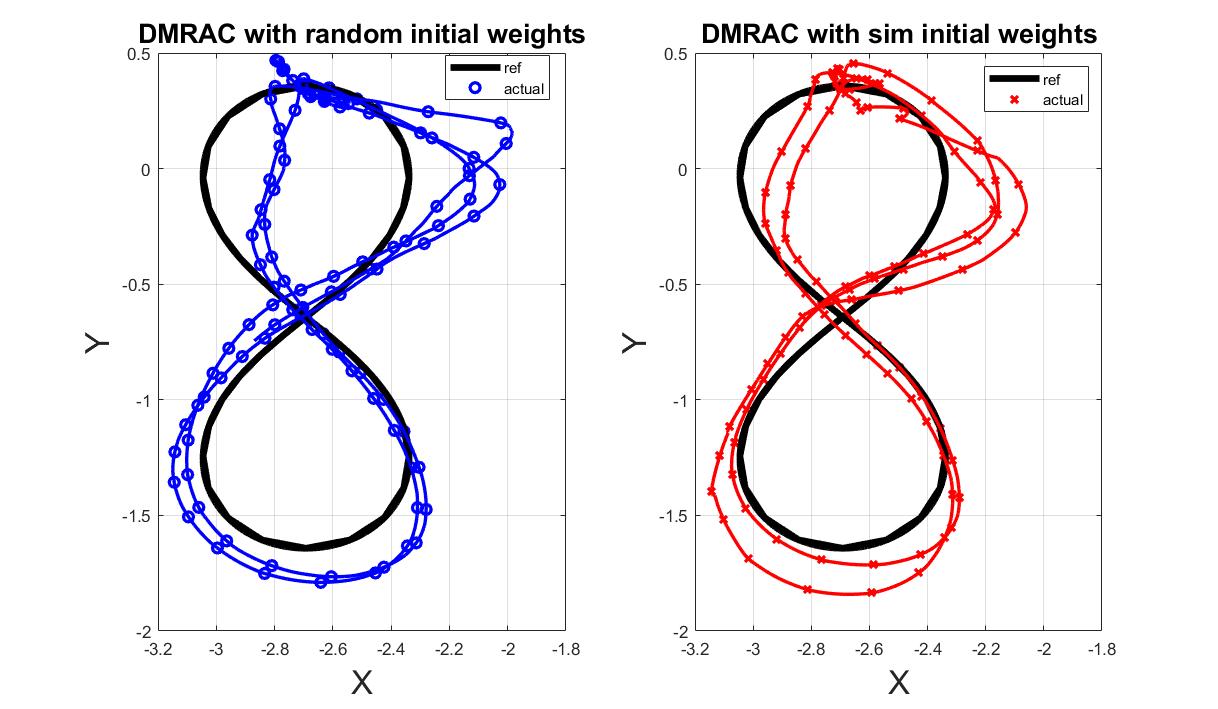}
    \caption{Figure of 8 Trajectory tracking under wind bias with random initialization vs. Feature transfer from simulation to Real in DMRAC}
    \label{fig:DMRAC_sim}
\end{figure}

The following experiments are similar to one in the previous section. Here we are investigating the network representation transfer from simulation to the real world. In this experiment, DMRAC is run in a simulation environment, where the network is trained over data collected through the simulated drone follow a figure of 8 trajectories without any disturbance. These trained network weights are then used as initialization weights for the case where DMRAC is experimented on the actual physical quadrotor. The controller performance is compared between the randomly initialized DMRAC vs. DMRAC initialized with network weights from the simulated quadrotor. We test the two controllers performing a figure of 8 trajectory maneuvers under high wind bias. Figure-\ref{fig:DMRAC_sim} shows the improvement in DMRAC's performance when the initial weights are from simulation rather than being initialized entirely randomly.

\subsubsection{Learning Retention}
\begin{figure}[tbh!]
    \centering
    \includegraphics[width=0.8\textwidth]{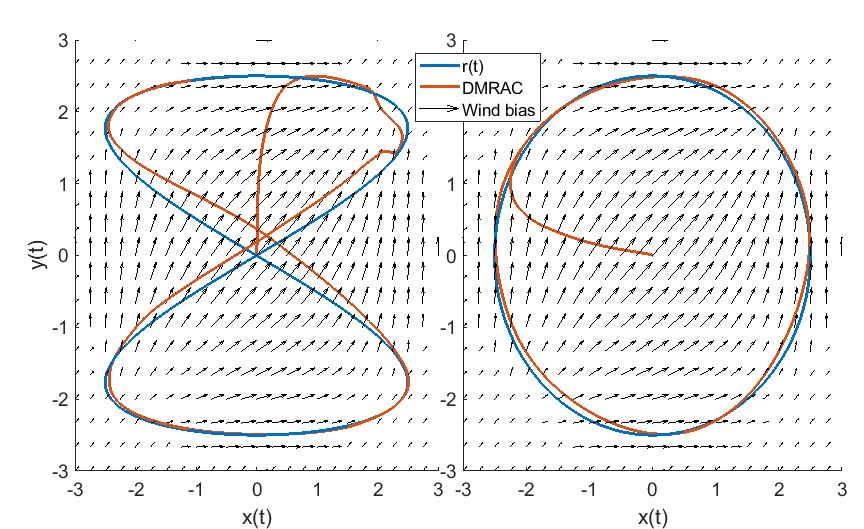}
    \caption{DMRAC Controller Evaluation on 6DOF Quadrotor dynamics model: DMRAC Controllers on quadrotor trajectory tracking with active learning for figure-8 tracking and DMRAC as frozen feed-forward network (Circular Trajectory) to test network generalization}
    \label{fig:LR_fig1}
\end{figure}

\begin{figure}[tbh!]
    \centering
    \includegraphics[width=0.8\textwidth]{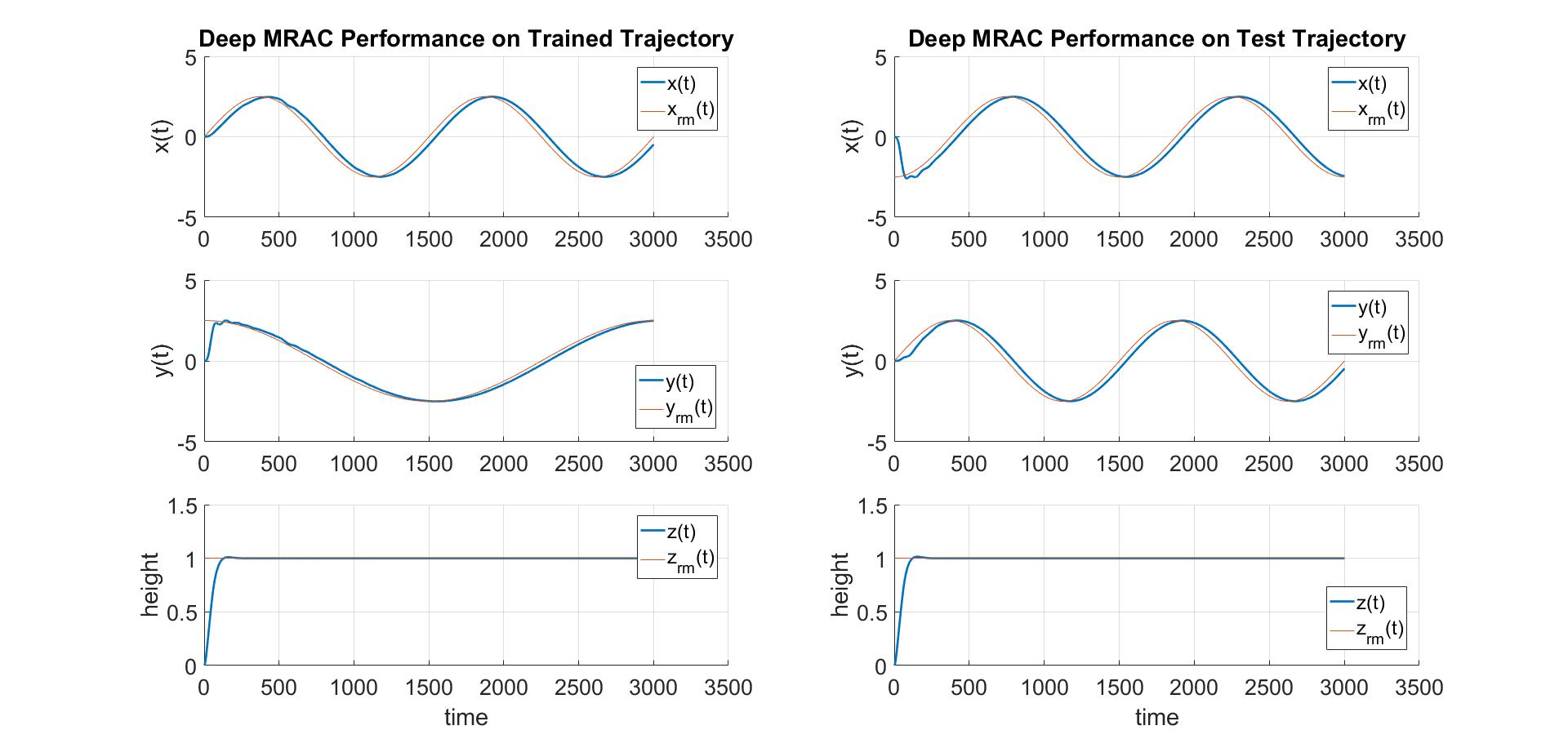}
    \caption{DMRAC Controller Evaluation on 6DOF Quadrotor dynamics model: Closed-loop system tracking performance in position $x, y$ and height $z$}
    \label{fig:LR_fig2}
\end{figure}

\begin{figure}[tbh!]
    \centering
    \includegraphics[width=0.8\textwidth]{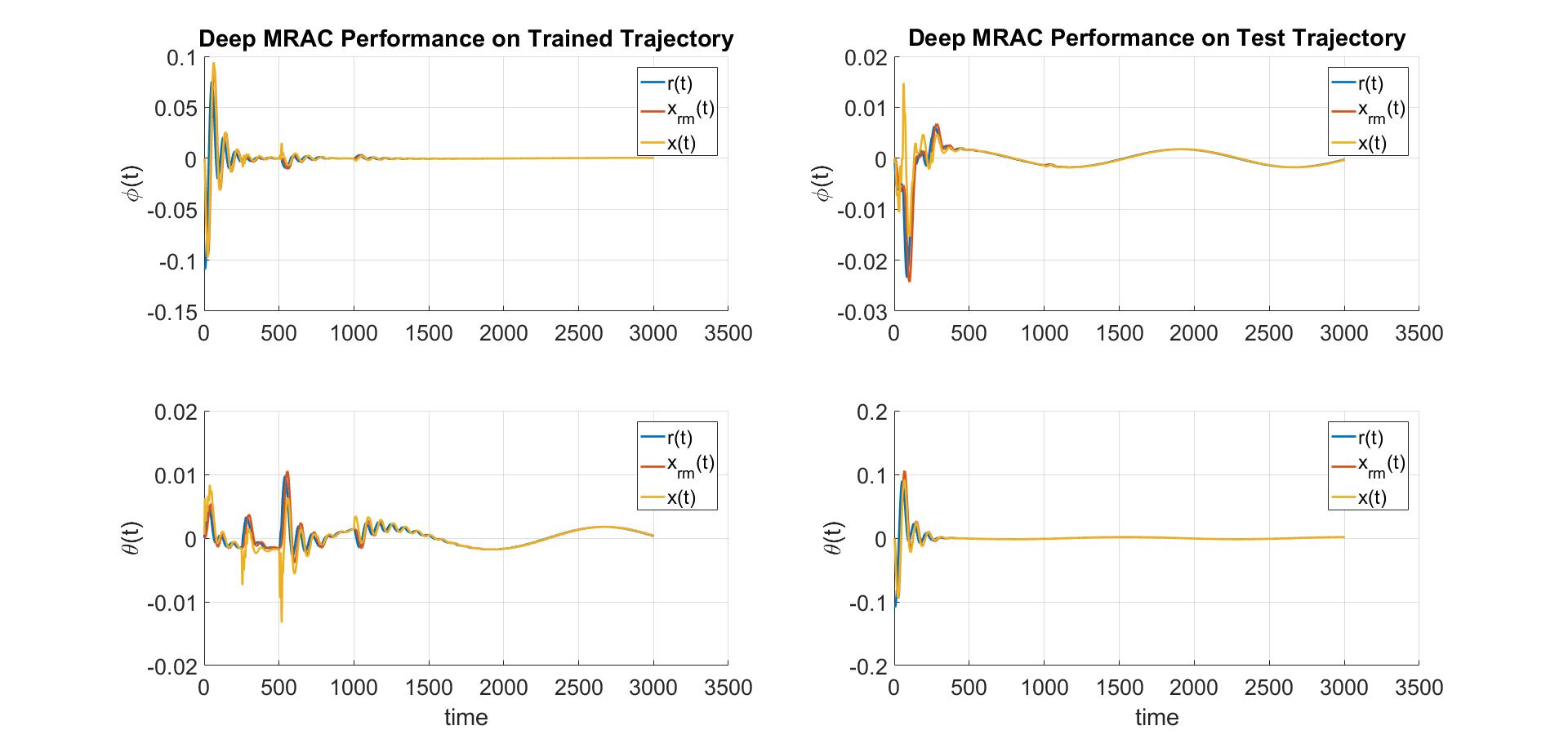}
    \caption{DMRAC Controller Evaluation on 6DOF Quadrotor dynamics model: Closed-loop system tracking performance in roll rate $\phi(t)$ and Pitch $\theta(t)$}
    \label{fig:LR_fig3}
\end{figure}

\begin{figure}[tbh!]
    \centering
    \includegraphics[width=0.6\textwidth]{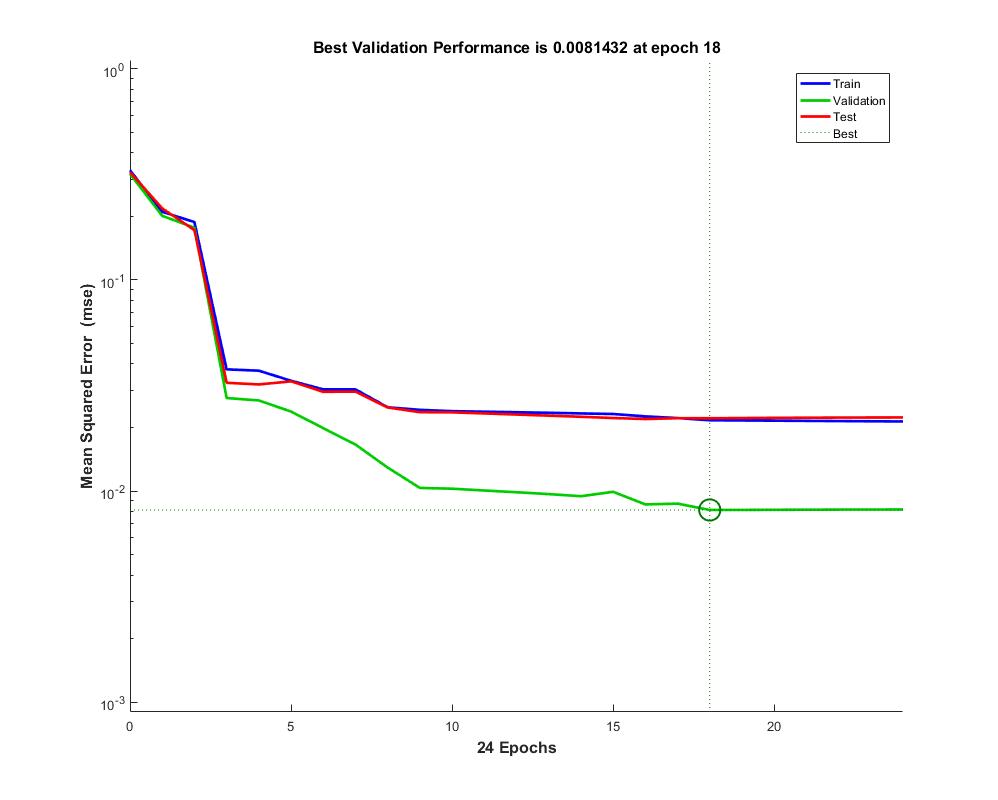}
    \caption{DMRAC Controller Evaluation on 6DOF Quadrotor dynamics model: Deep neural network training performance over train, test and validation sets.}
    \label{fig:LR_fig4}
\end{figure}

Figure-\ref{fig:LR_fig1},\ref{fig:LR_fig2} and Fig-\ref{fig:LR_fig3} show the closed loop system performance in tracking the reference signal for DMRAC controller and learning retention when used as the feed-forward network on a similar trajectory (Circular) with no learning. We demonstrate the proposed DMRAC controller under uncertainty and without domain information is successful in producing desired reference tracking. Since DMRAC, unlike traditional MRAC, uses DNN for uncertainty estimation is hence capable of retaining the past learning and thereby can be used in tasks with similar features without active online adaptation Figure-\ref{fig:LR_fig2}-\ref{fig:LR_fig3}. Whereas traditional MRAC which is ``pointwise in time" learning algorithm and cannot generalize across tasks. The presented controller achieves tighter tracking with smaller tracking error in both outer and inner loop states as shown in Fig-\ref{fig:LR_fig2} and Fig-\ref{fig:LR_fig3} in both with adaptation and as a feed-forward adaptive network without adaptation.  Figure-\ref{fig:LR_fig4} demonstrate the DNN learning performance vs epochs. The Training, Testing and Validation error over the data buffer for DNN, demonstrate the network performance in learning a model of the system uncertainties and its generalization capabilities over unseen test data. 

\end{document}